\documentclass[11pt,a4paper]{article}
\usepackage[hyperref]{emnlp2020}
\usepackage{times}
\usepackage{latexsym}

\usepackage{microtype}
\usepackage{amsmath}
\usepackage{amssymb}

\usepackage{graphicx}
\usepackage{comment}
\usepackage{array}
\usepackage{mathtools}
\usepackage[linguistics]{forest}
\usepackage{xcolor}
\usepackage{comment}

\aclfinalcopy 

\makeatletter
\newcommand{\printfnsymbol}[1]{%
  \textsuperscript{\@fnsymbol{#1}}%
}
\makeatother

\title{Improving Aspect-based Sentiment Analysis with Gated Graph Convolutional Networks and Syntax-based Regulation}

\author{Amir Pouran Ben Veyseh\textsuperscript{\rm 1}\thanks{\text{ } Equal contribution.}, \text{ } Nasim Nouri\printfnsymbol{1}, Franck Dernoncourt\textsuperscript{\rm 2},\\
{\bf Quan Hung Tran}\textsuperscript{\rm 2}, {\bf Dejing Dou}\textsuperscript{\rm 1} and {\bf Thien Huu Nguyen}\textsuperscript{\rm 1,3} \\
\textsuperscript{\rm 1} Department of Computer and Information Science, University of Oregon,
\\Eugene, OR 97403, USA\\
\textsuperscript{\rm 2} Adobe Research, San Jose, CA, USA\\
\textsuperscript{\rm 3} VinAI Research, Vietnam\\
  \texttt{\{apouranb,dou,thien\}@cs.uoregon.edu}, \\ {\tt nasim.nourii@gmail.com}, \texttt{\{dernonco,qtran\}@adobe.com}
}

\date{}

\begin{document}
\maketitle
\begin{abstract}
Aspect-based Sentiment Analysis (ABSA) seeks to predict the sentiment polarity of a sentence toward a specific aspect. Recently, it has been shown that dependency trees can be integrated into deep learning models to produce the state-of-the-art performance for ABSA. However, these models tend to compute the hidden/representation vectors without considering the aspect terms and fail to benefit from the overall contextual importance scores of the words that can be obtained from the dependency tree for ABSA. In this work, we propose a novel graph-based deep learning model to overcome these two issues of the prior work on ABSA. In our model, gate vectors are generated from the representation vectors of the aspect terms to customize the hidden vectors of the graph-based models toward the aspect terms. In addition, we propose a mechanism to obtain the importance scores for each word in the sentences based on the dependency trees that are then injected into the model to improve the representation vectors for ABSA. The proposed model achieves the state-of-the-art performance on three benchmark datasets.

\end{abstract}

\section{Introduction}

Aspect-based Sentiment Analysis (ABSA) is a fine-grained version of sentiment analysis (SA) that aims to find the sentiment polarity of the input sentences toward a given aspect. We focus on the term-based aspects for ABSA where the aspects correspond to some terms (i.e., sequences of words) in the input sentence. For instance, an ABSA system should be able to return the negative sentiment for input sentence ``\textit{The staff were very polite, but the quality of the food was terrible.}'' assuming ``{\it food}'' as the aspect term.

Due to its important applications (e.g., for opinion mining), ABSA has been studied extensively in the literature. In these studies, deep learning has been employed to produce the state-of-the-art performance for this problem \cite{tang:16-effective,Ma:17}. Recently, in order to further improve the performance, the syntactic dependency trees have been integrated into the deep learning models \cite{huang:19,zhang:19-aspect} for ABSA (called the graph-based deep learning models). Among others, dependency trees help to directly link the aspect term to the syntactically related words in the sentence, thus facilitating the graph convolutional neural networks (GCN) \cite{kipf:17} to enrich the representation vectors for the aspect terms.


However, there are at least two major issues in these graph-based models that should be addressed to boost the performance. First, the representation vectors for the words in different layers of the current graph-based models for ABSA are not customized for the aspect terms. This might lead to suboptimal representation vectors where the irrelevant information for ABSA might be retained and affect the model's performance. Ideally, we expect that the representation vectors in the deep learning models for ABSA should mainly involve the related information for the aspect terms, the most important words in the sentences. Consequently, in this work, we propose to regulate the hidden vectors of the graph-based models for ABSA using the information from the aspect terms, thereby filtering the irrelevant information for the terms and customizing the representation vectors for ABSA. In particular, we compute a gate vector for each layer of the graph-based model for ABSA leveraging the representation vectors of the aspect terms. This layer-wise gate vector would be then applied over the hidden vectors of the current layer to produce customized hidden vectors for ABSA. In addition, we propose a novel mechanism to explicitly increase the contextual distinction among the gates to further improve the representation vectors.


The second limitation of the current graph-based deep learning models is the failure to explicitly exploit the overall importance of the words in the sentences that can be estimated from the dependency trees for the ABSA problem. In particular, a motivation of the graph-based models for ABSA is that the neighbor words of the aspect terms in the dependency trees would be more important for the sentiment of the terms than the other words in the sentence. The current graph-based models would then just focus on those syntactic neighbor words to induce the representations for the aspect terms. However, based on this idea of important words, we can also assign a score for each word in the sentences that explicitly quantify its importance/contribution for the sentiment prediction of the aspect terms. In this work, we hypothesize that these overall importance scores from the dependency trees might also provide useful knowledge to improve the representation vectors of the graph-based models for ABSA. Consequently, we propose to inject the knowledge from these syntax-based importance scores into the graph-based models for ABSA via the consistency with the model-based importance scores. In particular, using the representation vectors from the graph-based models, we compute a second score for each word in the sentences to reflect the model's perspective on the importance of the word for the sentiment of the aspect terms. The syntax-based importance scores are then employed to supervise the model-based importance scores, serving as a method to introduce the syntactic information into the model. In order to compute the model-based importance scores, we exploit the intuition that a word would be more important for ABSA if it is more similar the overall representation vector to predict the sentiment for the sentence in the final step of the model. In the experiments, we demonstrate the effectiveness of the proposed model with the state-of-the-art performance on three benchmark datasets for ABSA. In summary, our contributions include:

$\bullet$ A novel method to regulate the GCN-based representation vectors of the words using the given aspect term for ABSA.

$\bullet$ A novel method to encourage the consistency between the syntax-based and model-based importance scores of the words based on the given aspect term.

$\bullet$ Extensive experiments on three benchmark datasets for ABSA, resulting in new state-of-the-art performance for all the datasets.

\section{Related Work}

Sentiment analysis has been studied under different settings in the literature (e.g., sentence-level, aspect-level, cross-domain) \cite{wang:19,zhang:19-tree,sun:19,chauhan:19,hu:19}. For ABSA, the early works have performed feature engineering to produce useful features for the statistical classification models (e.g., SVM) \cite{kiritchenko:14}. Recently, deep learning models have superseded the feature based models due to their ability to automatically learn effective features from data \cite{tang:16-effective,johnson:15,tang:16-aspect}. The typical network architectures for ABSA in the literature involve convolutional neural networks (CNN) \cite{johnson:15}, recurrent neural networks (RNN) \cite{tang:16-effective}, memory networks \cite{tang:16-aspect}, attention \cite{luong:15} and gating mechanisms \cite{he:18}. The current state-of-the-art deep learning models for ABSA feature the graph-based models where the dependency trees are leveraged to improve the performance. \cite{huang:19,zhang:19-aspect,hou:19}. However, to the best of our knowledge, none of these works has used the information from the aspect term to filter the graph-based hidden vectors and exploited importance scores for words from dependency trees as we do in this work.

\section{Model}

The task of ABSA can be formalized as follows: Given a sentence $X=[x_1,x_2,\ldots,x_n]$ of $n$ words/tokens and the index $t$ ($1\leq t \leq n$) for the aspect term $x_t$, the goal is to predict the sentiment polarity $y^*$ toward the aspect term $x_t$ for $X$. Our model for ABSA in this work consists of three major components: (i) Representation Learning, (ii) Graph Convolution and Regulation, and (iii) Syntax and Model Consistency.

{\bf (i) Representation Learning}: Following the recent work in ABSA \cite{huang:19,song:19}, we first utilize the contextualized word embeddings BERT \cite{Devlin:19} to obtain the representation vectors for the words in $X$. In particular, we first generate a sequence of words of the form $\hat{X} = [CLS] + X + [SEP] + x_t + [SEP]$ where $[CLS]$ and $[SEP]$ are the special tokens in BERT. This word sequence is then fed into the pre-trained BERT model to obtain the hidden vectors in the last layer. Afterwards, we obtain the embedding vector $e_i$ for each word $x_i \in X$ by averaging the hidden vectors of $x_i$’s sub-word units (i.e., wordpiece). As the result, the input sentence $X$ will be represented by the vector sequence $E=e_1,e_2,\ldots,e_n$ in our model. Finally, we also employ the hidden vector $s$ for the special token $[CLS]$ in $\hat{X}$ from BERT to encode the overall input sentence $X$ and its aspect term $x_t$.

\label{sec:graph-gate}
{\bf (ii) Graph Convolution and Regulation}: In order to employ the dependency trees for ABSA, we apply the GCN model \cite{Nguyen:18,Veyseh:19b} to perform $L$ abstraction layers over the word representation vector sequence $E$. A hidden vector for a word $x_i$ in the current layer of GCN is obtained by aggregating the hidden vectors of the dependency-based neighbor words of $x_i$ in the previous layer. Formally, let $h^l_i$ ($0 \le l \le L$, $1 \le i \le n$) be the hidden vector of the word $x_i$ at the $l$-th layer of GCN. At the beginning, the GCN hidden vector $h^0_i$ at the zero layer will be set to the word representation vector $e_i$. Afterwards, $h^l_i$ ($l > 0$) will be computed by: $h^l_i = ReLU(W_l\hat{h}^l_i)$, $\hat{h}^l_i = \Sigma_{j\in N(i)} h^{l-1}_j / |N(i)|$ where $N(i)$ is the set of the neighbor words of $x_i$ in the dependency tree. We omit the biases in the equations for simplicity.

One problem with the GCN hidden vectors $h^l_i$ GCN is that they are computed without being aware of the aspect term $x_t$. This might retain irrelevant or confusing information in the representation vectors (e.g., a sentence might have two aspect terms with different sentiment polarity). In order to explicitly regulate the hidden vectors in GCN to focus on the provided aspect term $x_i$, our proposal is to compute a gate $g_l$ for each layer $l$ of GCN using the representation vector $e_t$ of the aspect term: $g_l = \sigma(W^g_l e_t)$. This gate is then applied over the hidden vectors $h^l_i$ of the $l$-th layer via the element-wise multiplication $\circ$, generating the regulated hidden vector $\bar{h}^l_i$ for $h^l_i$: $\bar{h}^l_i = g_l \circ h^l_i$.


Ideally, we expect that the hidden vectors of GCN at different layers would capture different levels of contextual information in the sentence. The gate vectors $g_t$ for these layer should thus also exhibit some difference level for contextual information to match those in the GCN hidden vectors. In order to explicitly enforce the gate diversity in the model, our intuition is to ensure that the regulated GCN hidden vectors, once obtained by {\it applying different gates to the same GCN hidden vectors}, should be distinctive. This allows us to exploit the contextual information in the hidden vectors of GCN to ground the information in the gate vectors for the explicit gate diversity promotion. 

In particular, given the $l$-th layer of GCN, we first obtain an overall representation vector $\bar{h}^l$ for the regulated hidden vectors at the $l$-th layer using the max-pooled vector: $\bar{h}^l = max\_pool(\bar{h}^l_1,\ldots, \bar{h}^l_n)$. Afterwards, we apply the gate vectors $g^{l'}$ from the other layers ($l' \ne l$) to the GCN hidden vectors $h^l_i$ at the $l$-th layer, resulting in the regulated hidden vectors $\bar{h}^{l,l'}_i = g^{l'} \circ h^l_i$. For each of these other layers $l'$, we also compute an overall representation vector $\bar{h}^{l,l'}$ with max-pooling: $\bar{h}^{l,l'} = max\_pool(\bar{h}^{l,l'}_1, \ldots, \bar{h}^{l,l'}_n)$. Finally, we promote the diversity between the gate vectors $g^l$ by enforcing the distinction between $\bar{h}^l$ and $\bar{h}^{l,l'}$ for $l' \ne l$. This can be done by minimizing the cosine similarity between these vectors, leading to the following regularization term $L_{div}$ to be added to the loss function of the model:
$$\mathcal{L}_{div} = \frac{1}{L(L-1)}\Sigma_{l=1}^L\Sigma_{l'=1,l'\neq l}^L \bar{h}^l \cdot \bar{h}^{l,l'}$$.

{\bf (iii) Syntax and Model Consistency}: As presented in the introduction, we would like to obtain the importance scores of the words based on the dependency tree of $X$, and then inject these syntax-based scores into the graph-based deep learning model for ABSA to improve the quality of the representation vectors. Motivated by the contextual importance of the neighbor words of the aspect terms for ABSA, we use the negative of the length of the path from $x_i$ to $x_t$ in the dependency tree to represent the syntax-based importance score $syn_i$ for $x_i \in X$. For convenience, we also normalize the scores $syn_i$ with the softmax function.


In order to incorporate syntax-based scores $syn_i$ into the model, we first leverage the hidden vectors in GCN to compute a model-based importance score $mod_i$ for each word $x_i \in X$ (also normalized with softmax). Afterwards, we seek to minimize the KL divergence between the syntax-based scores $syn_1,\ldots,syn_n$ and the model-based scores $mod_1,\ldots,mod_n$ by introducing the following term $L_{const}$ into the overall loss function: $L_{const} = -syn_i \log \frac{syn_i}{mod_i}$. The rationale is to promote the consistency between the syntax-based and model-based importance scores to facilitate the injection of the knowledge in the syntax-based scores into the representation vectors of the model.



For the model-based importance scores, we first obtain an overall representation vector $V$ for the input sentence $X$ to predict the sentiment for $x_t$. In this work, we compute $V$ using the sentence representation vector $s$ from BERT and the regulated hidden vectors in the last layer of GCN: $V = [s, max\_pool(\hat{h}^L_1,\ldots,\hat{h}^L_n)]$. Based on this overall representation vector $V$, we consider a word $x_i$ to be more contextually important for ABSA than the others if its regulated GCN hidden vector $\hat{h}^L_i$ in the last GCN layer is more similar to $V$ than those for the other words. The intuition is the GCN hidden vector of a contextually important word for ABSA should be able capture the necessary information to predict the sentiment for $x_t$, thereby being similar to $V$ that is supposed to encode the overall relevant context information of $X$ to perform sentiment classification. In order to implement this idea, we use the dot product of the transformed vectors for $V$ and $\hat{h}^L_i$ to determine the model-based importance score for $x_i$ in the model: $mod_i = \sigma(W_V V)\cdot \sigma(W_H \hat{h}^L_i)$.

Finally, we feed $V$ into a feed-forward neural network with softmax in the end to estimate the probability distribution $P(.|X,x_t)$ over the sentiments for $X$ and $x_t$. The negative log-likelihood $L_{pred} = -\log P(y^*|X,x_t)$ is then used as the prediction loss in this work. The overall loss to train the proposed model is then: $\mathcal{L} =  \mathcal{L}_{div} + \alpha \mathcal{L}_{const} + \beta \mathcal{L}_{pred}$ where $\alpha$ and $\beta$ are trade-off parameters.

\section{Experiments}

\hspace{0.33cm} {\bf Datasets and Parameters}: We employ three datasets to evaluate the models in this work. Two datasets, Restaurant and Laptop, are adopted from the SemEval 2014 Task 4 \cite{pontiki:14} while the third dataset, MAMS, is introduced in \cite{jiang:19}. All the three datasets involve three sentiment categories, i.e., positive, neural, and negative. The numbers of examples for different portions of the three datasets are shown in Table \ref{tab:statistics}.

\begin{table}[ht]
    \centering
\resizebox{.35\textwidth}{!}{ 
   \begin{tabular}{l|c|c|c}
        \hline
       \textbf{Dataset}  & \textbf{Pos.} & \textbf{Neu.} & \textbf{Neg.} \\ \hline
        Restaurant-Train & 2164 & 637 & 807 \\
        Restaurant-Test & 728 & 196 & 196 \\
        Laptop-Train & 994 & 464 & 870 \\
        Laptop-Test & 341 & 169 & 128 \\
        MAMS-Train & 3380 & 5042 & 2764 \\
        MAMS-Dev & 403 & 604 & 325 \\
        MAMS-Test & 400 & 607 & 329 \\
    \end{tabular}
}
    \caption{Statistics of the datasets}
    \label{tab:statistics}
\end{table}

As only the MAMS dataset provides the development data, we fine-tune the model's hyper-parameters on the development data of MAMS and use the same hyper-parameters for the other datasets. The following hyper-parameters are suggested for the proposed model by the fine-tuning process: 200 dimensions for the hidden vectors of the feed forward networks and GCN layers, 2 hidden layers in GCN, the size 32 for the mini-batches, the learning rate of 0.001 for the Adam optimizer, and 1.0 for the trade-off parameters $\alpha$ and $\beta$. Finally, we use the cased BERT$_{base}$ model with 768 hidden dimensions in this work.

{\bf Results}: To demonstrate the effectiveness of the proposed method, we compare it with the following baselines: (1) the feature-based model that applies feature engineering and the SVM model \cite{kiritchenko:14}, (2) the deep learning models based on the sequential order of the words in the sentences, including CNN, LSTM, attention and the gating mechanism \cite{tang:16-effective,wang:16,tang:16-aspect,huang:18,jiang:19}, and (3) the graph-based models that exploit dependency trees to improve the deep learning models for ABSA \cite{huang:19,zhang:19-aspect,hou:19,sun:19,Wang:20}.

Table \ref{tab:results} presents the performance of the models on the test sets of the three benchmark datasets. This table shows that the proposed model outperforms all the baselines over different benchmark datasets. The performance gaps are significant with $p < 0.01$, thereby demonstrating the effectiveness of the proposed model for ABSA.

\begin{table}[ht] 
\centering 
\resizebox{.49\textwidth}{!}{ 
\begin{tabular}{l|cc|cc|cc} 
\hline 
\textbf{Model} & \multicolumn{2}{c}{\textbf{Rest.}} & \multicolumn{2}{c}{\textbf{Laptop}} & \multicolumn{2}{c}{\textbf{MAMS}} \\
& Acc. & F1 & Acc. & F1 & Acc. & F1 \\ \hline
 SVM \shortcite{kiritchenko:14} & 80.2 & - & 70.5 & - & - & - \\ \hline 
 TD-LSTM \shortcite{tang:16-effective} & 78.0 & 66.7 & 68.8 & 68.4 & 74.6 & - \\ 
 AT-LSTM \shortcite{wang:16} & 76.2 & - & 68.9 & - & 77.6 & - \\ 
 MemNet \shortcite{tang:16-aspect} & 79.6 & 69.6 & 70.6 & 65.1 & 64.6 & - \\ 
 AOA-LSTM \shortcite{huang:18} & 79.9 & 70.4 & 72.6 & 67.5 & 77.3 & - \\ 
 CapsNet \shortcite{jiang:19} & 80.7 & - & - & - & 79.8 & - \\ \hline 
 ASGCN \shortcite{zhang:19-aspect} & 80.8 & 72.1 & 75.5 & 69.2 & - & - \\ 
 GAT \shortcite{huang:19} & 81.2 & - & 74.0 & - & - & - \\ 
 CDT \shortcite{sun:19} & 82.3 & 74.02 & 77.1 & 72.9 & - & - \\ 
 R-GAT \shortcite{Wang:20} & 83.3 & 76.0 & 77.4 & 73.7 & - & - \\ 
 \hline BERT* \shortcite{jiang:19} & 84.4 & - & 77.1 & - & 82.2 & - \\ 
 GAT* \shortcite{huang:19} & 83.0 & - & 80.1 & - & - & - \\ 
 AEN-BERT* \shortcite{song:19} & 84.4 & 76.9 & 79.9 & 76.3 & - & - \\ 
 SA-GCN* \shortcite{hou:19} & 85.8 & 79.7 & 81.7 & 78.8 & - & - \\ 
 CapsNet* \shortcite{jiang:19} & 85.9 & - & - & - & 83.4 & - \\
 R-GAT* \shortcite{Wang:20} & 86.6 & 81.3 & 78.2 & 74.0 & - & - \\ \hline 
 \textbf{The proposed model} & \textbf{87.2} & \textbf{82.5} & \textbf{82.8} & \textbf{80.2} & \textbf{88.2} & \textbf{57.1} \\
 \end{tabular} 
 } 
 \caption{Accuracy and F1 scores of the models on the test sets. * indicates the models with BERT.} \label{tab:results} 
 \end{table}


{\bf Ablation Study}: There are three major components in the proposed model: (1) the gate vectors $g_l$ to regulate the hidden vectors of GCN (called {\bf Gate}), (2) the gate diversity component $L_{div}$ to promote the distinction between the gates (called {\bf Div.}), and (3) the syntax and model consistency component $L_{const}$ to introduce the knowledge from the syntax-based importance scores (called {\bf Con.}). Table \ref{tab:ablation} reports the performance on the MAMS development set for the ablation study  when the components mentioned in each row are removed from the proposed model. Note that the exclusion of {\bf Gate} would also remove {\bf Div.} due to their dependency. It is clear from the table that all three components are necessary for the proposed model as removing any of them would hurt the model's performance.




\begin{table}[ht]
    \centering
    \small
    \begin{tabular}{l|c|c}
        \textbf{Model} & \textbf{Acc}. & \textbf{F1} \\ \hline
        \textbf{The proposed model (full)} & \textbf{87.98} & \textbf{57.2} \\ \hline
        -Div. & 87.33 & 56.6 \\
        -Con. & 86.82 & 56.2 \\
        -Div -Con. & 86.52 & 56.1 \\
        -Gate & 86.25 & 55.8 \\
        -Gate -Con. & 86.02 & 54.3
    \end{tabular}
    \caption{Ablation study on MAMS dev set}
    \label{tab:ablation}
\end{table}

{\bf Gate Diversity Analysis}: In order to enforce the diversity of the gate vectors $g_t$ for different layers of GCN, the proposed model indirectly minimizes the cosine similarities between the regulated hidden vectors of GCN at different layers (i.e., in $L_{div}$). The regulated hidden vectors are obtained by applying the gate vectors to the hidden vectors of GCN, serving as a method to ground the information in the gates with the contextual information in the input sentences (i.e., via the hidden vectors of GCN) for diversity promotion. In order to demonstrate the effectiveness of such gate-context grounding mechanism for the diversity of the gates, we evaluate a more straightforward baseline where the gate diversity is achieved by directly minimizing the cosine similarities between the gate vectors $g_t$ for different GCN layers. In particular, the diversity loss term $L_{div}$ in this baseline would be: $\mathcal{L}_{div} = \frac{1}{L(L-1)}\Sigma_{l=1}^L\Sigma_{l'=1,l'\neq l}^L g_l \cdot g_{l'}$. We call this baseline {\bf GateDiv.} for convenience. Table \ref{tab:div-gate} report the performance of GateDiv. and the proposed model on the development dataset of MAMS. As can be seen, the proposed model is significantly better than GateDiv., thereby testifying to the effectiveness of the proposed gate diversity component with information-grounding in this work. We attribute this superiority to the fact that the regulated hidden vectors of GCN provide richer contextual information for the diversity term $L_{div}$ than those with the gate vectors. This offers better grounds to support the gate similarity comparison in $L_{div}$, leading to the improved performance for the proposed model.


\begin{table}[ht]
    \centering
\resizebox{.28\textwidth}{!}{
    \begin{tabular}{l|c}
        \textbf{Model} & \textbf{Acc}. \\ \hline
        \textbf{The proposed model} & \textbf{87.98} \\ \hline
        GateDiv. & 86.13 \\
    \end{tabular}
}
    \caption{Model performance on the MAMS development set when the diversity term $L_{div}$ is directly computed from the gate vectors.}
    \label{tab:div-gate}
\end{table}

\section{Conclusion}

We introduce a new model for ABSA that addresses two limitations of the prior work. It employs the given aspect terms to customize the hidden vectors. It also benefits from the overall dependency-based importance scores of the words. Our extensive experiments on three benchmark datasets empirically demonstrate the effectiveness of the proposed approach, leading to state-of-the-art results on these datasets. The future work involves applying the proposed model to the related tasks for ABSA, e.g., event detection \cite{Nguyen:15b}.




\section*{Acknowledgement}


This research is based upon work supported in part by the Office of the Director of National Intelligence (ODNI), Intelligence Advanced Research Projects Activity (IARPA), via IARPA Contract No. 2019-19051600006 under the Better Extraction from Text Towards Enhanced Retrieval (BETTER) Program. The views and conclusions contained herein are those of the authors and should not be interpreted as necessarily representing the official policies, either expressed or implied, of ODNI, IARPA, the Department of Defense, or the U.S. Government. The U.S. Government is authorized to reproduce and distribute reprints for governmental purposes notwithstanding any copyright annotation therein. This document does not contain technology or technical data controlled under either the U.S. International Traffic in Arms Regulations or the U.S. Export Administration Regulations.

\bibliography{emnlp2020}

\begin{thebibliography}{26}
\expandafter\ifx\csname natexlab\endcsname\relax\def\natexlab#1{#1}\fi

\bibitem[{Chauhan et~al.(2019)Chauhan, Akhtar, Ekbal, and
  Bhattacharyya}]{chauhan:19}
Dushyant~Singh Chauhan, Md~Shad Akhtar, Asif Ekbal, and Pushpak Bhattacharyya.
  2019.
\newblock Context-aware interactive attention for multi-modal sentiment and
  emotion analysis.
\newblock In \emph{EMNLP}.

\bibitem[{Dehong et~al.(2017)Dehong, Sujian, Xiaodong, and Houfeng}]{Ma:17}
Ma~Dehong, Li~Sujian, Zhang Xiaodong, and Wang Houfeng. 2017.
\newblock Interactive attention networks for aspect-level sentiment
  classification.
\newblock In \emph{IJCAI}.

\bibitem[{Devlin et~al.(2019)Devlin, Chang, Lee, and Toutanova}]{Devlin:19}
Jacob Devlin, Ming-Wei Chang, Kenton Lee, and Kristina Toutanova. 2019.
\newblock {BERT}: Pre-training of deep bidirectional transformers for language
  understanding.
\newblock In \emph{NAACL-HLT}.

\bibitem[{He et~al.(2018)He, Lee, Ng, and Dahlmeier}]{he:18}
Ruidan He, Wee~Sun Lee, Hwee~Tou Ng, and Daniel Dahlmeier. 2018.
\newblock Effective attention modeling for aspect-level sentiment
  classification.
\newblock In \emph{ACL}.

\bibitem[{Hou et~al.(2019)Hou, Huang, Wang, Huang, He, and Zhou}]{hou:19}
Xiaochen Hou, Jing Huang, Guangtao Wang, Kevin Huang, Xiaodong He, and Bowen
  Zhou. 2019.
\newblock Selective attention based graph convolutional networks for
  aspect-level sentiment classification.
\newblock In \emph{arXiv}.

\bibitem[{Hu et~al.(2019)Hu, Wu, Zhao, Guo, Cheng, and Su}]{hu:19}
Mengting Hu, Yike Wu, Shiwan Zhao, Honglei Guo, Renhong Cheng, and Zhong Su.
  2019.
\newblock Domain-invariant feature distillation for cross-domain sentiment
  classification.
\newblock In \emph{EMNLP}.

\bibitem[{Huang and Carley(2019)}]{huang:19}
Binxuan Huang and Kathleen Carley. 2019.
\newblock Syntax-aware aspect level sentiment classification with graph
  attention networks.
\newblock In \emph{EMNLP}.

\bibitem[{Huang et~al.(2018)Huang, Ou, and Carley}]{huang:18}
Binxuan Huang, Yanglan Ou, and Kathleen~M Carley. 2018.
\newblock Aspect level sentiment classification with attention-over-attention
  neural networks.
\newblock In \emph{SBP-BRiMS}.

\bibitem[{Jiang et~al.(2019)Jiang, Chen, Xu, Ao, and Yang}]{jiang:19}
Qingnan Jiang, Lei Chen, Ruifeng Xu, Xiang Ao, and Min Yang. 2019.
\newblock A challenge dataset and effective models for aspect-based sentiment
  analysis.
\newblock In \emph{ACL}.

\bibitem[{Johnson and Zhang(2015)}]{johnson:15}
Rie Johnson and Tong Zhang. 2015.
\newblock Semi-supervised convolutional neural networks for text categorization
  via region embedding.
\newblock In \emph{NIPS}.

\bibitem[{Kipf and Welling(2017)}]{kipf:17}
Thomas~N. Kipf and Max Welling. 2017.
\newblock Semi-supervised classification with graph convolutional networks.
\newblock In \emph{ICLR}.

\bibitem[{Luong et~al.(2015)Luong, Pham, and Manning}]{luong:15}
Thang Luong, Hieu Pham, and Christopher~D. Manning. 2015.
\newblock Effective approaches to attention-based neural machine translation.
\newblock In \emph{EMNLP}.

\bibitem[{Nguyen and Grishman(2015)}]{Nguyen:15b}
Thien~Huu Nguyen and Ralph Grishman. 2015.
\newblock Event detection and domain adaptation with convolutional neural
  networks.
\newblock In \emph{ACL}.

\bibitem[{Nguyen and Grishman(2018)}]{Nguyen:18}
Thien~Huu Nguyen and Ralph Grishman. 2018.
\newblock Graph convolutional networks with argument-aware pooling for event
  detection.
\newblock In \emph{AAAI}.

\bibitem[{Pontiki et~al.(2014)Pontiki, Galanis, Pavlopoulos, Papageorgiou,
  Androutsopoulos, and Manandhar}]{pontiki:14}
Maria Pontiki, Dimitris Galanis, John Pavlopoulos, Harris Papageorgiou, Ion
  Androutsopoulos, and Suresh Manandhar. 2014.
\newblock {S}em{E}val-2014 task 4: Aspect based sentiment analysis.
\newblock In \emph{SemEval}.

\bibitem[{Song et~al.(2019)Song, Wang, Jiang, Liu, and Rao}]{song:19}
Youwei Song, Jiahai Wang, Tao Jiang, Zhiyue Liu, and Yanghui Rao. 2019.
\newblock Attentional encoder network for targeted sentiment classification.
\newblock In \emph{arXiv}.

\bibitem[{Sun et~al.(2019)Sun, Zhang, Mensah, Mao, and Liu}]{sun:19}
Kai Sun, Richong Zhang, Samuel Mensah, Yongyi Mao, and Xudong Liu. 2019.
\newblock Aspect-level sentiment analysis via convolution over dependency tree.
\newblock In \emph{EMNLP}.

\bibitem[{Tang et~al.(2016)Tang, Qin, and Liu}]{tang:16-aspect}
Duyu Tang, Bing Qin, and Ting Liu. 2016.
\newblock Aspect level sentiment classification with deep memory network.
\newblock In \emph{EMNLP}.

\bibitem[{Veyseh et~al.(2019)Veyseh, Nguyen, and Dou}]{Veyseh:19b}
Amir Pouran~Ben Veyseh, Thien~Huu Nguyen, and Dejing Dou. 2019.
\newblock Graph based neural networks for event factuality prediction using
  syntactic and semantic structures.
\newblock In \emph{ACL}.

\bibitem[{Wagner et~al.(2014)Wagner, Arora, Cortes, Barman, Bogdanova, Foster,
  and Tounsi}]{kiritchenko:14}
Joachim Wagner, Piyush Arora, Santiago Cortes, Utsab Barman, Dasha Bogdanova,
  Jennifer Foster, and Lamia Tounsi. 2014.
\newblock {NRC}-canada-2014: Detecting aspects and sentiment in customer
  reviews.
\newblock In \emph{SemEval}.

\bibitem[{Wagner et~al.(2016)Wagner, Arora, Cortes, Barman, Bogdanova, Foster,
  and Tounsi}]{tang:16-effective}
Joachim Wagner, Piyush Arora, Santiago Cortes, Utsab Barman, Dasha Bogdanova,
  Jennifer Foster, and Lamia Tounsi. 2016.
\newblock Effective {LSTM}s for target-dependent sentiment classification.
\newblock In \emph{COLING}.

\bibitem[{Wang et~al.(2019)Wang, Sun, Li, Wang, Si, Zhang, Liu, and
  Zhou}]{wang:19}
Jingjing Wang, Changlong Sun, Shoushan Li, Jiancheng Wang, Luo Si, Min Zhang,
  Xiaozhong Liu, and Guodong Zhou. 2019.
\newblock Human-like decision making: Document-level aspect sentiment
  classification via hierarchical reinforcement learning.
\newblock In \emph{EMNLP}.

\bibitem[{Wang et~al.(2020)Wang, Shen, Yang, Quan, and Wang}]{Wang:20}
Kai Wang, Weizhou Shen, Yunyi Yang, Xiaojun Quan, and Rui Wang. 2020.
\newblock Relational graph attention network for aspect-based sentiment
  analysis.
\newblock In \emph{ACL}.

\bibitem[{Wang et~al.(2016)Wang, Huang, Zhu, and Zhao}]{wang:16}
Yequan Wang, Minlie Huang, Xiaoyan Zhu, and Li~Zhao. 2016.
\newblock Attention-based {LSTM} for aspect-level sentiment classification.
\newblock In \emph{EMNLP}.

\bibitem[{Zhang et~al.(2019)Zhang, Li, and Song}]{zhang:19-aspect}
Chen Zhang, Qiuchi Li, and Dawei Song. 2019.
\newblock Aspect-based sentiment classification with aspect-specific graph
  convolutional networks.
\newblock In \emph{EMNLP}.

\bibitem[{Zhang and Zhang(2019)}]{zhang:19-tree}
Yuan Zhang and Yue Zhang. 2019.
\newblock Tree communication models for sentiment analysis.
\newblock In \emph{ACL}.

\end{thebibliography}
\bibliographystyle{acl_natbib}

\clearpage

\appendix

\end{document}